\documentclass[letterpaper, 10 pt, conference]{ieeeconf}  

\IEEEoverridecommandlockouts                              

\pdfoutput=1
\usepackage{graphics} 
\usepackage{epsfig} 
\usepackage{mathptmx} 
\usepackage{times} 
\usepackage{amsmath} 
\usepackage{amssymb}  
\usepackage{subcaption}
\usepackage{hyperref}
\usepackage{makecell}

\title{\LARGE \bf
Bring Your Own Grasp Generator: Leveraging Robot Grasp Generation for Prosthetic Grasping
}

\author{Giuseppe Stracquadanio$^{1}$, Federico Vasile$^{1,2}$, Elisa Maiettini$^{1}$, Nicolò Boccardo$^{3}$ and Lorenzo Natale$^{1}$
\thanks{$^{1}$Giuseppe Stracquadanio, Federico Vasile, Elisa Maiettini and Lorenzo Natale are with the Istituto Italiano di Tecnologia, Humanoid Sensing and Perception, 16163 Genoa, Italy (email: \texttt{name}.\texttt{surname}@iit.it).}%
\thanks{$^{2}$Federico Vasile is also with the Dipartimento di Informatica, Bioingegneria, Robotica e Ingegneria dei Sistemi (DIBRIS), University of Genova, 16146 Genoa, Italy.}%
\thanks{$^{3}$Nicolò Boccardo is with the Istituto Italiano di Tecnologia, Rehab Technologies, 16163 Genoa, 
Italy, and also with the Open University Affiliated Research Centre at Istituto Italiano di Tecnologia (ARC@IIT), 16163 Genoa, Italy (email: nicolo.boccardo@iit.it) .
       }%
\thanks{This work received support by the European Union’s Horizon-JU-SNS-2022 Research and Innovation Programme under the project TrialsNet (Grant Agreement No. 101095871) and the project RAISE (Robotics and AI for Socio-economic Empowerment) implemented under the National Recovery and Resilience Plan, Mission 4 funded by the European Union – NextGenerationEU.}
}

\begin{document}

\maketitle
\thispagestyle{empty}
\pagestyle{empty}

\begin{abstract}
One of the most important research challenges in upper-limb prosthetics is enhancing the user-prosthesis communication to closely resemble the experience of a natural limb. As prosthetic devices become more complex, users often struggle to control the additional degrees of freedom. In this context, leveraging \textit{shared-autonomy} principles can significantly improve the usability of these systems. 
In this paper, we present a novel \textit{eye-in-hand} prosthetic grasping system that follows these principles. Our system initiates the approach-to-grasp action based on user's command and automatically configures the DoFs of a prosthetic hand. First, it reconstructs the 3D geometry of the target object without the need of a depth camera. Then, it tracks the hand motion during the approach-to-grasp action and finally selects a candidate grasp configuration according to user's intentions. We deploy our system on the Hannes prosthetic hand and test it on able-bodied subjects and amputees to validate its effectiveness.  We compare it with a multi-DoF prosthetic control baseline and find that our method enables faster grasps, while simplifying the user experience. Code and demo videos are available online at \href{https://hsp-iit.github.io/byogg/}{this https URL}.

\end{abstract}

\section{INTRODUCTION}
\begin{figure}[h]
    \centering
    \includegraphics[width=\linewidth]{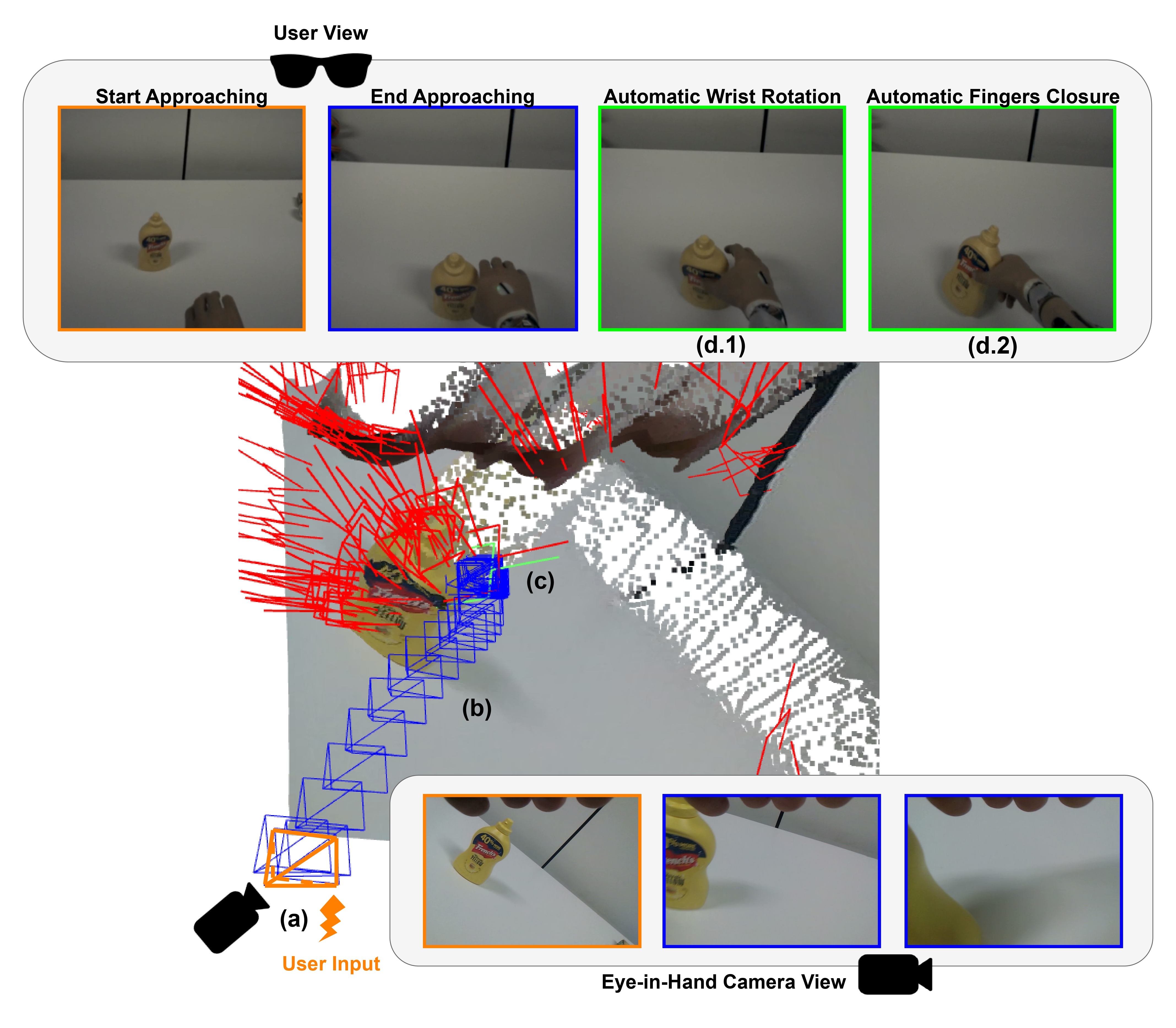}
    \caption{The phases of our grasping pipeline. A demonstration of semi-autonomous grasp is shown through the user perspective.}
    \label{fig:pipeline-high-level}
    \vspace{-0.6cm}
\end{figure}

Upper limb amputation can drastically change the quality of life of people, impacting their ability to carry out actions that seemed trivial.
Recovering the functionality of the amputated limb becomes, up to a certain extent, possible with prostheses. Modern upper-limb prosthetic devices try to push this extent, seeking a seamless integration and embodiment with the user and trying to replicate the key properties of a real human hand. 
Currently, commercial upper-limb prostheses are based on \textit{electromyography} (EMG) or \textit{mechanomyography} (MMG) as user-input interfaces, allowing users to control the Degrees of Freedom (DoFs) of the prosthetic device by relating the user signals to motors' velocities \cite{marinelli2023active}. A standard approach to multi-DoF control consists in the Sequential Switching and Control (SSC) paradigm. Following this method, the user drives each joint individually through EMG signals and an explicit trigger is required to switch between them. However, as the number of available DoFs increases, the control becomes complex and unintuitive \cite{amsuess2014}.
For this reason, simplifying the user-prosthesis communication is one of the most interesting research problems in prosthetics. An emerging research direction proposes a \textit{shared-autonomy} \cite{seminara2023hierarchical,marinelli2023active} framework, in which the collaboration with a semi-autonomous system relieves the users from exhausting and unnatural actions, while still being able to operate the device according to their intentions. This framework usually leverages additional input modalities, such as inertial measurements, images and depth information to accomplish the task \cite{starke2022semi,castro2022continuous,vasile2022grasp,hundhausen2023grasping}. 
For instance, previous works have already considered the integration of a RGB camera for visual recognition to predict a grasp pre-shape \cite{starke2022semi, castro2022continuous,vasile2022grasp}. Following such approaches, in this work, we introduce a novel eye-in-hand vision-based prosthetic grasping pipeline, 
drawing inspiration from 
the human cognitive process involved in the action of grasping.
Generally, when we observe an object that we want to grasp, we perceive its geometry and, unconsciously, evaluate the most suitable grasp among different possibilities. Immediately, we also begin to plan the trajectory that our hand will follow to reach the object, and finally grasp it. 
Each module of our semi-autonomous control system is designed to emulate the steps of this process.
First of all, we use a depth estimation network to reconstruct geometric information about the object. Then, we exploit a grasp pose generation model to predict multiple grasp candidates (Fig.~\ref{fig:pipeline-high-level}a). Subsequently, a visual odometry model estimates the hand trajectory during the approach (Fig.~\ref{fig:pipeline-high-level}b), allowing us to select a candidate according to the user's intentions  (Fig.~\ref{fig:pipeline-high-level}c). Finally, we execute this pose on our prosthesis  (Fig.~\ref{fig:pipeline-high-level}d).
We make the following contributions: (a) we propose a novel eye-in-hand prosthetic grasping pipeline, designed to improve the user experience by automatically setup a grasp configuration according to user's intentions, (b) we deploy it on Hannes \cite{laffranchi2020hannes}, validating its effectiveness on able-bodied subjects and (c) testing its robustness and embodiment on amputees, while conducting (d) an early-stage analysis on cognitive load.

\section{RELATED WORK}
\subsection{Monocular Depth Estimation}
Monocular depth estimation is the task of estimating the depth for each pixel of a single RGB image. Knowing the depth, together with the camera calibration matrix $K$, allows to un-project each image pixel back to 3D.
Current state-of-the-art methods are mostly based on transformer foundational models, such as DINOv2 \cite{oquabdinov2} or Depth-Anything \cite{yang2024depth}. To allow generalization across multiple datasets, these methods are trained with an affine-invariant depth estimation objective \cite{ranftl2020towards}, regressing a \textit{relative} (up-to-scale) depth. 
However, some applications, like ours, would require a \textit{metric} (i.e., with \textit{absolute} scale) depth. In this case, zero-shot inference to a new scenario is not directly possible, before proper fine-tuning on a custom dataset. Therefore, in this work, we generate a synthetic dataset to fine-tune our model. 

\subsection{Vision-based Prosthetic Grasping}
Several vision-driven prosthetic grasping pipelines have been proposed in literature. 
In \cite{starke2022semi}, the authors propose an \textit{eye-in-hand} system based on visual recognition of a \textit{known} target object and sensor-fusion to guide the selection of the candidate grasp trajectory from an existing database. The automatic execution of a grasp pre-shape was studied by \cite{shi2022target}, simultaneously predicting the target object in clutter and the current step in the temporal evolution of the grasping action. To support different grasp types for a single object, \cite{vasile2022grasp} proposes a synthetic data generation pipeline leveraging the approaching trajectory to automatically label the object parts with grasp pre-shapes \cite{laffranchi2020hannes,boccardo2023development}. 
A pipeline based on depth-perception from a depth camera is described in \cite{castro2022continuous}, where a grasp size and wrist rotation were estimated after fitting geometric primitives to the object.
Recently, \cite{hundhausen2023grasping} proposed to reconstruct the object geometry to infer information (i.e., the object diameter) for automatic finger closure, relying on a time-of-flight depth sensor. 
In this work, similarly to \cite{castro2022continuous} and \cite{hundhausen2023grasping}, we rely on geometric structure of the objects, but employ a monocular depth estimation network and avoid the need for any depth sensor. 
Moreover, differently from previous works, we aim to leverage the know-how from robotic grasping methods and apply it to a prosthetic scenario.
Specifically, grasping hypotheses are computed at the beginning of the action, when the full object is visible in the camera frustum and are subsequently tracked during the action using a visual odometry module. 
This allows detecting the most suitable grasping candidate from the initial hypothesis and finally execute it on the prosthetic hand.

\subsection{Vision-based Robotic Grasping}
Two different research directions can be identified in the vision-based robotic grasping literature, depending on the type and DoFs of the robot end-effector. The first, dealing with parallel-jaw grippers, includes works that present several gripper parameterizations and end-to-end architectures for learning to \textit{generate} \cite{mousavian20196,murali20206} or \textit{regress} \cite{sundermeyer2021contact,alliegro2022end} grasp distributions over point clouds. 
The second direction, instead, studies grasping with humanoid and dexterous hands \cite{ferreira2020unigrasp,Karunratanakul2020grasping,kelin2022efficient,li2023gendexgrasp}. Although these methods allow a higher diversity of dexterous grasps, we decided to build on top of the current state-of-the-art with grippers that, in addition to being more mature, has an immediate simplicity of use without making particular hypotheses on hand kinematics. 
In this work, we adapt the Contact-GraspNet \cite{sundermeyer2021contact} gripper representation to our multi-DoFs Hannes hand \cite{boccardo2023development}. Other similar works, such as \cite{alliegro2022end}, can be adapted with little modifications to account for a different gripper parameterization. 

\subsection{Visual Odometry}
Visual Odometry (VO) is the task of estimating camera position and orientation using visual information.
A closely related problem is Simultaneous Localization and Mapping (SLAM), where a stream of images is processed in real time to estimate the camera position and simultaneously build a 3D map of the environment \cite{mur2015orb,teed2021droid,teed2024deep}. Both VO and SLAM typically use feature tracking at the pixel level to establish 2D-2D correspondences across consecutive frames, which are then used as inputs for optimization objectives, to solve for the camera poses and 3D point coordinates. SLAM methods further continuously compute corrections to improve map global consistency and mitigate the drift resulting from accumulated errors in VO. 
We rely on a SLAM method to perform VO, in order to benefit from this additional robustness. 
More specifically, we integrate DPVO \cite{teed2024deep} to process streams of images in real time. This method performs sparse RGB patch tracking. This increases inference speed and lowers the memory usage, making its adoption suitable in a prosthetic scenario. However, it is fair to say that any other real-time VO technique can be easily adapted into our grasping architecture.

\section{METHODS}
\begin{figure*}[!ht]
    \centering
    \vspace{+0.3cm}
    \includegraphics[width=0.95\textwidth]{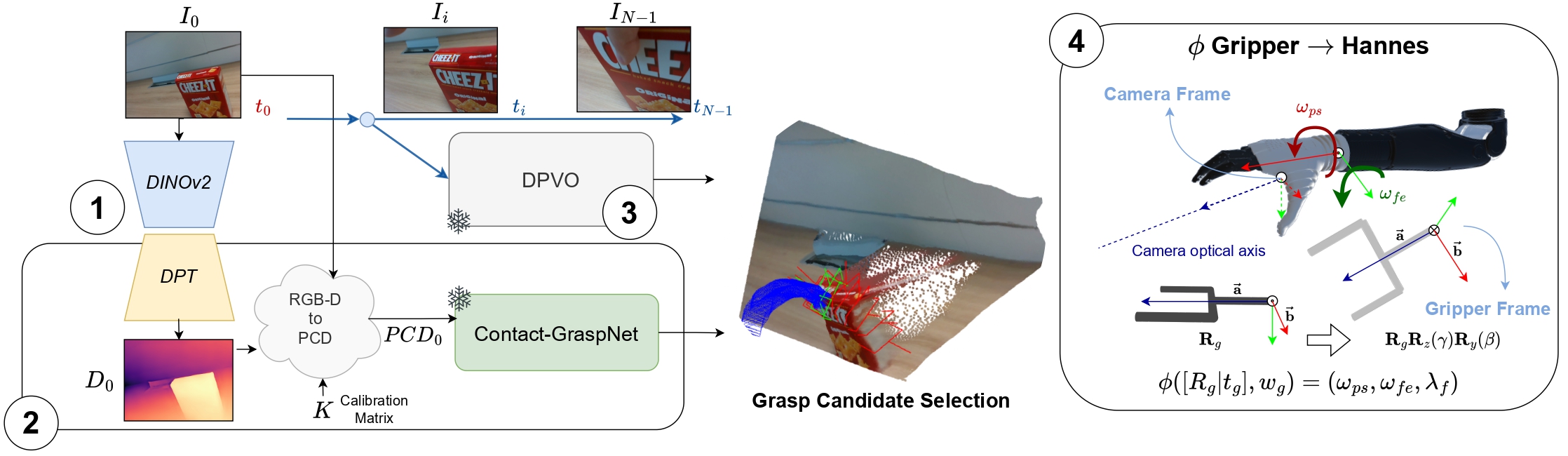}
    \caption{Details of our grasping pipeline.   (1) A depth $D_0$ is estimated from the first frame of the grasping sequence. (2)  $D_0$ is used to build the point cloud $PCD_0$ and generate a distribution of grasp poses. (3) The hand-trajectory is estimated by a visual odometry module, and used to select a grasp candidate. (4) The candidate pose is mapped to Hannes.}
    \label{fig:pipeline-architecture}
    \vspace{-0.6cm}
\end{figure*}

\subsection{Overview}
We designed our vision-based prosthetic grasping system following the shared-autonomy framework. Specifically, the user is responsible for pointing their hand at the target object and triggering the starting signal, using the EMG user-interface. 
Then, the user approaches the object, while the system performs online tracking of the hand position and finally predicts the most proper Hannes configuration for grasping.
More precisely, as soon as the starting signal is triggered, an image from the eye-in-hand camera is captured and grasp candidates are predicted (Fig.~\ref{fig:pipeline-architecture}.2). However, since the grasp generation method requires a point cloud as input, we first estimate a depth map (Fig.~\ref{fig:pipeline-architecture}.1) and reconstruct a point cloud. Then, given the grasp candidates in the scene, the most suitable one is selected according to the approach-to-grasp trajectory performed by the user. To this end, during the approach, the visual odometry module continuously processes the RGB images from the eye-in-hand camera and estimates the camera trajectory in space (Fig.~\ref{fig:pipeline-architecture}.3). Finally, the closest grasp candidate to the camera is selected and executed on the Hannes prosthesis (Fig.~\ref{fig:pipeline-architecture}.4).

\subsection{Architecture Details}
\subsubsection{\textbf{Monocular Depth Estimation}}
We rely on a model with a DPT \cite{ranftl2021vision} head and a DINOv2 \cite{oquabdinov2} backbone for monocular depth estimation. 
Given the first RGB frame of the grasping sequence $\mathbf{I}_0 \in \mathbb{R}^{H\times W}$, we estimate a depth-map $\mathbf{D}_0 \in \mathbb{R}^{H\times W}$, where each value $d_{u,v}, u \in [0,...,H-1], v \in [0,...,W-1]$ is the depth in absolute scale (i.e., \textit{metric} depth) estimated for the corresponding pixel in $\mathbf{I}_0$. 

\subsubsection{\textbf{Grasp Generation}}
Given $\mathbf{I}_0$, $\mathbf{D}_0$ and the intrinsic camera  parameters $K$, we build our point-cloud $\mathbf{PCD}_0$ 
by computing the 3D coordinates $(x,y,z)$ for each pixel $(u,v) \in \mathbf{I}_0$ in the camera coordinate system.
We use Contact-GraspNet \cite{sundermeyer2021contact} to process the resulting point cloud.
Contact-GraspNet associates each gripper pose to a visible contact point $\mathbf{c}$, reducing the complexity of the learning problem. 
More specifically, the $\mathbb{SE}(3)$ position of a gripper is parameterized using two orthonormal vectors, the \textit{approach} vector $\mathbf{a} \in \mathbb{R}^3$ and the \textit{grasp baseline} vector $\mathbf{b} \in \mathbb{R}^3$, intersecting on the contact point $\mathbf{c}$.
Furthermore, to reduce the memory requirements, the point cloud is first downsampled to $n=20k$ points.  
In order to increase the number of grasps generated on object contact points, Contact-GraspNet can rely on object segmentation masks and sequentially process different point-cloud segments. 
We decided to not rely on additional object segmentation models to avoid introducing biases to known objects. We also experimented with the same unknown object segmentation model \cite{xie2021unseen} used in~\cite{sundermeyer2021contact} and observed lower performance in our scenario. 
We modify the Contact-GraspNet downsampling strategy in order to assign a higher sampling probability to closer points to the \textit{eye-in-hand} camera, usually corresponding to contact points on the target object, and lower probabilities to farther points, commonly associated to background (e.g., a wall). For doing that, we sample pixels from a distribution having the probability scores 
$p_{u,v} = softmax(1/d_{u,v}),\ \forall (u,v) \in \mathbf{I}_0$
, where $d_{u,v}$ is the estimated depth for pixel at location $(u,v)$ and the softmax is computed over all the pixels.



\subsubsection{\textbf{Visual Odometry}}
 We use DPVO \cite{teed2024deep} to estimate the position and orientation of the Hannes hand, using the sequence of RGB frames collected from the eye-in-hand camera while approaching the target object.
DPVO performs sparse patch-tracking
to build 2D-2D correspondences between consecutive frames. 
Each patch $\mathbf{P}_k$ is represented as the homogeneous array $\mathbf{P}_k = (\mathbf{u}_k, \mathbf{v}_k, \mathbf{1}, \mathbf{d}_k),\ \mathbf{u}_k,\mathbf{v}_k,\mathbf{d}_k \in \mathbb{R}^{1\times p^2}$, containing the pixel coordinates and the depth of each pixel in a $p \times p$ patch. 
At each new incoming RGB frame, DPVO estimates the 2D motion of tracked patches and then solves for both camera poses $\mathbf{T}_j \in \mathbb{SE}(3)$ and patch representations $\mathbf{P}_k \in \mathbb{R}^{4 \times p^2}$.
Because DPVO works with RGB frames, the optimized camera poses $\mathbf{T}_j = [\mathbf{R}_j | \mathbf{t}_j]$ will have up-to-scale translation components. We compute a scaling factor by comparing the optimized patch representations $\mathbf{P}_k$ having the initial frame (i.e., $\mathbf{I}_0$) as \textit{source} frame with the \textit{absolute}-scale dense depth estimated by our monocular depth estimation model, $\mathbf{D}_0$. Specifically, as DPVO assumes the same depth value for every pixel in a patch, we consider the center pixel of each patch. Then, we sample from the dense depth map at the same coordinates, and compute the \textit{median} ratio between the two depth values for each sampled pixel. Formally, we compute the new camera poses $\mathbf{T}^{*}_{j} = [\mathbf{R}_j | \alpha^{*}\mathbf{t}_j]$, with $\alpha^{*} = \psi(\mathbf{D}_0, \{\mathbf{P}_{k|\mathbf{I}_0}\}_{k})$, 
where $\psi$ is the operator that takes the center coordinates $u_c, v_c$ of each patch $\mathbf{P}_k$, samples $\mathbf{D}_0$ at the same coordinates and computes the \textit{median} of the ratios $\{ d^{\mathbf{D}_0}_{u_c,v_c} / d^{\mathbf{P}_k}_{u_c,v_c} \}_{k|\mathbf{I}_0}$ ($k|\mathbf{I}_0$ means ``patch $k$ sampled from $\mathbf{I}_0$").
We use the last estimated camera position to select the nearest grasp pose. We compute a euclidean distance between the camera position 
and the middle point of each gripper. We select as the candidate grasp pose the one having the shortest distance to the camera, and thus, to the Hannes hand. 
Finally, if this distance is below a given threshold (e.g., $5$cm), DPVO stops running and the candidate grasp is automatically executed on the Hannes prosthesis.
\subsubsection{\textbf{Mapping grasp candidates to the Hannes hand}}
A Hannes pre-shape configuration is defined by the triplet $(\omega_{ps}, \omega_{fe}, \lambda_{f})$, where $\lambda_{f}$ is the Hannes opening in the fingers \textit{opening-closing} (FOC) range, and $(\omega_{ps}, \omega_{fe})$ are, respectively the wrist \textit{pronation-supination} (WPS) and \textit{flexion-extension} (WFE) angles.
Finding $\lambda_{f}$ from gripper parameters is straightforward, as it only requires to scale the gripper width $w_g$ to the Hannes FOC range. 
We also define the optimal  $(\omega_{ps}, \omega_{fe})$ to be the joint angles that make 
the eye-in-hand camera optical axis match the gripper approaching direction $\vec{\mathbf{a}}$, while the direction $\mathbf{f}$ in which the Hannes fingers close (in which the other four fingers close to reach the thumb) matches the gripper baseline vector $\vec{\mathbf{b}}$.
We first apply a rotation of $\gamma = -\pi/2$ around the $z$-axis and $\beta = -\pi/4$ around $y$-axis to the gripper pose, such that a gripper with identity pose resembles the Hannes hand in the \textit{home} position (Fig.~\ref{fig:pipeline-architecture}.4).
Thus, if $\mathbf{R}_g \in \mathbb{SO}(3)$ is the rotation component of the \textit{candidate} grasp pose, we computed the \textit{desired} camera pose to perform the grasp as
$\mathbf{R}_{c}^{des} = (\mathbf{R}_{c}^{N-1})^{T} \mathbf{R}_{g} \mathbf{R}_z(\gamma) \mathbf{R}_y(\beta)$,
where $\mathbf{R}_{c}^{N-1}$ embeds the rotation of the last estimated camera pose (i.e., at frame $N-1$, with $N$ the number of processed frames). 
The transpose of $\mathbf{R}_{c}^{N-1}$ is pre-multiplied to project a gripper pose into the new reference frame of the camera. 
Then, we minimize the $\mathbb{SO}(3)$ error $\mathbf{e} = (\theta r_x, \theta r_z)^T$, where $\theta\mathbf{r} = axisangle(\mathbf{R}_{c}^{des})$.
We project this error to the joint space, using 
the Jacobian in the end-effector frame and the control law $\mathbf{\dot{q}} =\lambda \big({}^c\mathbf{J}_{e}[\dot{\theta_x},\dot{\theta_z}](\mathbf{q})\big)^{\dag} \mathbf{e}$, where ${}^c\mathbf{J}_{e}(\mathbf{q}) = {}^c\mathbf{Ad}_{e} {}^e\mathbf{J}_{e}(\mathbf{q})$, 
${}^c\mathbf{Ad}_{e}$ is the adjoint matrix that converts Jacobian velocities expressed in the end-effector frame to the camera frame, ${}^c\mathbf{J}_{e}[\dot{\theta_x},\dot{\theta_z}]$ is the $\mathbb{R}^{2 \times 2}$ matrix built by extracting from the Jacobian the rows corresponding to the angular velocities around $x$ and $z$-axis, $(\cdot)^{\dag}$ is the Moore-Penrose pseudo-inverse of a matrix, and $\mathbf{q} \in \mathbb{R}^2$ is the  vector encoding the Hannes wrist joint angles, $\mathbf{q} = (\omega_{ps}, \omega_{fe})^T$. We run this optimization until the norm of the error $\mathbf{e}$ is smaller than an error threshold, or for a fixed amount of steps (for constant time). We use the final $(\omega_{ps}, \omega_{fe})$ angles to control the Hannes wrist in the joint space.

\section{EXPERIMENTAL SETUP}
\subsection{Deployment and Embodiment}
We tested and deployed our grasping pipeline on our Hannes prosthesis. A small eye-in-hand RGB camera is embedded into the prosthesis palm. User input is handled using two EMG sensors, placed on the forearm flexor and extensor muscles. We use a pre-defined and easily identifiable EMG signal to let the user trigger the start of the \textit{approaching} stage. All our experiments were conducted by using a flexor and extensor \textit{co-contraction} as the triggering starting signal. 
When triggering the signal, the target object should be visible on the eye-in-hand camera in order for grasp candidates to be generated on that object.
At this point, the user can approach the target object. The end of the approaching stage, and thus the start of the \textit{grasping} stage, is automatically detected based on distance between the estimated camera and nearest gripper position. At the start of the \textit{grasping} stage, we control the wrist position (\textit{open-loop} control) by specifying joint angles $(\omega_{ps}, \omega_{fe})$. After a few seconds ($t_{grasp}$), we also send the $\lambda_f$ command to the fingers motor. We set $t_{grasp} = 2s$, after observing that this time is enough to let the user wrap the hand around the object, before the fingers are automatically closed for performing the grasp.
\subsection{Subjects, Goal and Target Objects}
We performed a first validation of our system on 10 able-bodied subjects, measuring a grasp success rate (GSR) and the average time to grasp (ATG). After validating the approach, we tested it on 3 amputee subjects.
Moreover, for amputees, we also compared our method with a control based solely on EMG sensors using the SSC paradigm (now referred to as EMG-SSC). Evaluating this baseline was possible only with amputee subjects, because already familiar with similar control strategies. Performing the same trials with able-bodied subjects was impractical, as it turned out to be too complex for people that had no prior experience with the embodiment.
We used 5 objects to conduct our experiments. The 3D model of three objects was included in synthetic data used to train our depth estimation model (\textit{known} objects). The other two, or similar objects, were not included (\textit{unknown} objects). The other components of our pipeline are object-agnostic. Objects are shown in Fig.~\ref{fig:able-bodied-plots} and Fig.~\ref{fig:amputees-plots}. 
Every user (able-bodied or amputee) was asked to perform 6 grasps for each object, for a total of 30 trials for each subject. Amputees also performed the same number of trials with EMG-SSC. 
Finally, we conducted an experimental study on \textit{cognitive load}, using the pupil dilation as a fatigue measure to further compare our method with EMG-SSC. 
The study adhered to the standard of the Declaration of Helsinki and was approved by the CET - Liguria ethical committee (Protocol code: IIT\_REHAB\_HT01). 
We refer to the supplementary video for demonstrations on both amputee and able-bodied subjects.
\subsection{Hardware}
We run our modules on a single NVIDIA RTX 3080 GPU. On this GPU, the pipeline can run at $30$ Hz during the approaching stage. To record pupil dilation during our trials with subjects, we used Tobii Pro Glasses 3, a wearable eye-tracking device able to record absolute measures of pupil diameter (at $100$ Hz) and other gaze measurements. 

\section{RESULTS}
\begin{figure*}[!t]
    \vspace{+0.3cm}
    \centering
    \includegraphics[width=\textwidth]{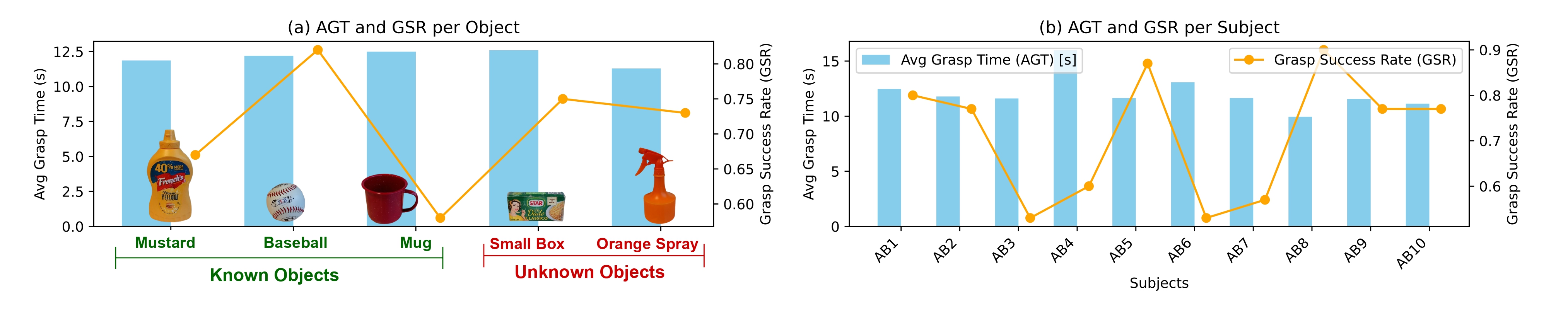}
    \caption{AGT and GSR measured on able-bodied subjects. Statistics are shown per object (a) and per subject (b).}
    \label{fig:able-bodied-plots}
    \vspace{-0.6cm}
\end{figure*}

\subsection{Ablation Study on Monocular Depth Estimation}
 To learn a monocular depth estimator, we use synthetic data to benefit from a significant amount of ground-truth depth values. We devised a synthetic data generation pipeline using the Unity engine and the Perception package \cite{unity-perception2022}. Our pipeline is based on \cite{vasile2022grasp} (\textit{SynV1}), but we extended it to specifically
increase the robustness of a depth estimation model. We propose several offline data augmentations (\textit{SynV2}). We simulate scenarios with clutter and occlusions by randomizing the number of objects in the scene. We randomize the camera optical axis direction to avoid biases to fixed object positions. 
Finally, we randomize light direction and intensity to perform renderings under variable light conditions.
To evaluate our depth estimation model, we use the same test data from \cite{vasile2022grasp}, collected using an Intel RealSense D435 depth camera. We compare two different backbone checkpoints (DINOv2 \cite{oquabdinov2} and DepthAnything \cite{yang2024depth}). We also experimented with two initialization strategies for the decoder head: random initialization and a checkpoint trained on the NYU-DepthV2 \cite{SilbermanECCV12} depth dataset. The encoder features were frozen (*) in some of our experiments. When fine-tuning the encoder features, we use the same recipe of DepthAnything. Results shown in Table~\ref{tab:depth-estimation-results} demonstrate that our \textit{SynV2} synthetic dataset is more effective for depth estimation and that random initialization of the decoder, with a fine-tuned encoder, produces the best results. 
\begin{table}[h!]
\centering
\caption{MDE evaluation on real test-set \cite{vasile2022grasp}}
\begin{tabular}{lcccc}
\Xhline{2\arrayrulewidth}
\textbf{Dataset} & \textbf{Encoder} & \textbf{Decoder} & \cite{Ladicky2014CVPR} $\delta_1$ $\uparrow$  & RMSE $\downarrow$ \\ \hline
\textit{SynV2} & DepthAnything &  Random & 0.689 & 0.115 \\
\textit{SynV2} & DINOv2 &  Random & \textbf{0.698} & \textbf{0.113}  \\ 
\textit{SynV2} & DINOv2\ (*) & NYUv2 & 0.658 & 0.166 \\
\textit{SynV1} & DINOv2\ (*) & NYUv2 & 0.412 & 0.476 \\
\Xhline{2\arrayrulewidth}
\end{tabular}
\label{tab:depth-estimation-results}
\vspace{-0.4cm}
\end{table}

\subsection{Experiments with Able-bodied Subjects}
Results for grasp success rate (GSR) and average grasp time (AGT) obtained on our 10 able-bodied subjects are shown in Fig.~\ref{fig:able-bodied-plots}. We report an analysis of GSR and AGT found on different objects (Fig.~\ref{fig:able-bodied-plots}a) and achieved by all the able-bodied subjects (Fig.~\ref{fig:able-bodied-plots}b). We can already extract useful information from this validation on able-bodied subjects. First, we can notice how our method does not exhibit generalization issues to unknown objects. We can also observe how the \textit{Mug} object was the most difficult object to grasp, achieving a GSR of $0.58$. 
Specifically, we were expecting a gripper to be predicted on the mug handle, or on the mug borders, such that the antipodal contact-points are placed, respectively, on the external and internal surface of the object. Instead, during our trials, most of the grippers were predicted with both contact-points on the external surface, with a large gripper aperture. However, even when fully-opened, our Hannes hand is not able to grasp the \textit{Mug} object this way, resulting in a failed attempt. Finally, we have evidence of a variable GSR on different able-bodied subjects ($0.71 \pm 0.14$), with AB3 and AB6 achieving only $0.53$ GSR, while AB5 and AB8 achieved $0.87$ and $0.90$ GSR, respectively. We believe a variable performance can be explained with a different subject response to the initial training stage or to muscle fatigue (e.g., due to holding the prosthesis for long periods of time). The AGT ($12.09 \pm 2.14$ over \textit{all} the trials) does not depend significantly on the object, as expected. Instead, subjects can approach the objects at different speeds or become familiar more quickly with the embodiment (especially with the EMG user-input interface). 
\subsection{Experiments with Amputees}
\begin{figure*}[t]
    \vspace{+0.3cm}
    \centering
    \includegraphics[width=\textwidth]{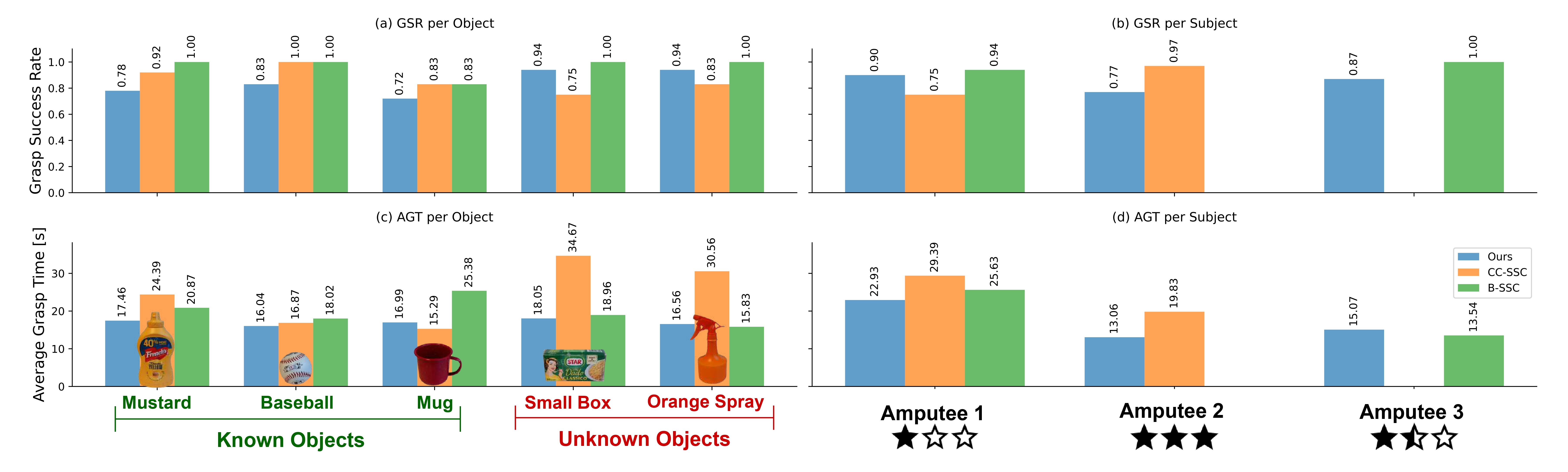}
    \caption{AGT and GSR measured on amputees. Results with our method are compared with CC-SSC and B-SSC baselines.}
    \label{fig:amputees-plots}
    \vspace{-0.3cm}
\end{figure*}

\begin{figure*}[t]
    \centering
    \includegraphics[width=\textwidth]{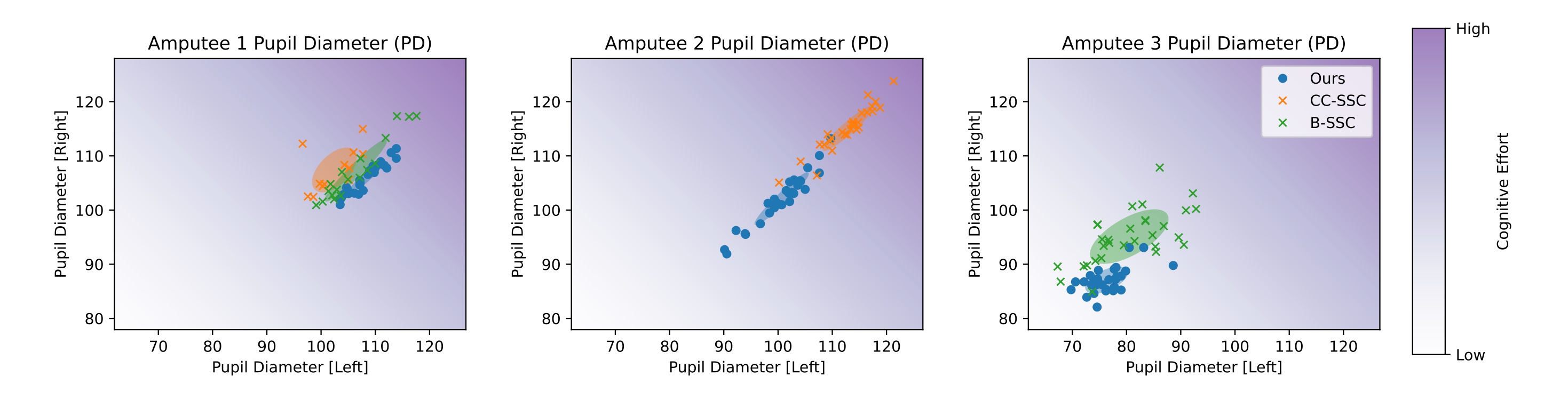}
    \caption{Distributions of pupil diameter (PD) recorded while performing trials.}
    \label{fig:cognitive-load-amputees}
    \vspace{-0.3cm}
\end{figure*}

We now report results and insights obtained from our analysis on our actual target users. Results for GSR and AGT are shown in Fig.~\ref{fig:amputees-plots}. 
We also report a comparison with results obtained for EMG-SSC baselines. We show a comparison summary in Table~\ref{tab:amputees-comparison}.
All our invited amputee subjects had already experience with the use of EMG sensors for prosthesis control. One of them (Amputee 2) also had experience with the same multi-DoF control paradigm used for our baselines. The level of prior experience and skill is indicated in Fig.~\ref{fig:amputees-plots} through the use of stars.
We experimented with a EMG-SSC baseline that uses a muscle co-contraction to trigger the joint switch (CC-SSC) and another EMG-SSC baseline in which the joint switch is operated through a button (B-SSC), held by the user using the other free hand.
We decided to use B-SSC to facilitate the continuation of the experiments, when we noticed that the amputees struggled to continuously use a co-contraction as a switch method, while having to use EMGs to control the joints of the prosthesis. 
Amputee 1 performed the first 40\% of the trials (12/30) with EMG-SSC. The remaining were performed using B-SSC. Amputee 2 performed all the trials with CC-SSC, and Amputee 3 performed all of them with B-SSC. 
Objects were picked in a random order. To avoid biasing our results, we never asked to grasp the same object for more consecutive trials for the EMG-SSC baselines. We were able to observe higher GSR results with amputee subjects, compared to able-bodied ones. This can be easily explained with a better response to fatigue over trials and with prior experience with the prosthetic embodiment. For EMG-SSC baselines, we expected nearly perfect GSR results with a trade-off on AGT. Interestingly, we measured a perfect GSR score with B-SSC on every object except the \textit{Mug}, confirming the intrinsic difficulty of grasping the object with our device. We also expected lower AGT times with B-SSC compared to CC-SSC, as the control is simplified at the cost of a more unhandy embodiment. We found that our method has, on average, lower AGT compared to both CC-SSC and B-SSC (Table~\ref{tab:amputees-comparison}). Notably, this is also true for a user who had already experience with the control baseline (Amputee 2), but no experience with our method (Fig.~\ref{fig:amputees-plots}d). This demonstrates that the proposed strategy enables faster grasps, while preserving the overall success rate.



\begin{table}[!t]
    \centering
    \caption{Summary table of results obtained on amputees}
    \begin{tabular}{lll}
    \Xhline{2\arrayrulewidth}
        \textbf{Method} & \textbf{$\uparrow$ \textbf{GSR}} & \textbf{$\downarrow$ \textbf{AGT} [s]} \\ \hline
        \textit{Ours} & 0.84 $\pm$ 0.12 & $\mathbf{17.02}$ $\pm$ $\mathbf{4.75}$\\ 
        CC-SSC & 0.87 $\pm$ 0.19 & 24.36 $\pm$ 15.07 \\ 
        B-SSC & \textit{0.97} $\pm$ \textit{0.15} & 19.81 $\pm$ 8.65 \\    
    \Xhline{2\arrayrulewidth}
    \end{tabular}
    \label{tab:amputees-comparison}
    \vspace{-0.4cm}
\end{table}

\subsection{Analysis on Cognitive Load}
\label{subsec:cognitive-load}
We conducted an analysis on cognitive load using eye-tracking data from Tobii Pro Glasses 3. Every amputee subject performed all the trials 
while wearing the eye-tracking glasses. For every subject, we also recorded a \textit{baseline} experiment, in which users were asked to grasp objects with their other real hand (from now on, \textit{Real-Hand}).  Following prior works doing the same analysis in similar scenarios \cite{cheng2023scoping,cognitive_load_biomechanics_2024}, we relate the pupil diameter (PD) to the user's cognitive load. Indeed, it has been observed that a dilation of pupil diameter is related to an increasing cognitive effort \cite{van2018pupil,waltercognitive2021}. Similarly to \cite{cognitive_load_biomechanics_2024}, for every user, we express PD as the percentage of the diameter measured during \textit{Real-Hand}. For our analysis, we only consider PD values on \textit{fixations}, to only include variations in PD which are actually due to cognitive load and not other factors, such as changes in gaze directions. Notice that our goal is not to present a statistical exhaustive study on the topic, but rather to demonstrate the applicability of this analysis, compared to other more obtrusive setups to evaluate cognitive load. 
We show amputees' PD plots in Fig.~\ref{fig:cognitive-load-amputees}. It is, indeed, interesting to observe how PD values are distributed for the three subjects. For Amputee 1, we do not observe significant differences in PD between the distributions. However, for Amputee 2 and 3, we observed, on average, an higher PD value while using a control baselines, compared to our method. Following these observations, we can hypothesize that our method was easier and more intuitive to use for Amputee 2 and 3, resulting in a lower measured cognitive effort.

\addtolength{\textheight}{-3cm}   
\section{CONCLUSIONS}
In this work, we introduced a novel vision-based prosthetic grasping pipeline, based on the shared-autonomy principles to enhance user-prosthesis interaction. 
Our system leverages components from the robotic literature and showcases how these can be applied in a prosthetic scenario.
We demonstrated the effectiveness of our framework by testing it on amputee subjects and comparing it with a standard multi-DoF control paradigm.
Finally, we performed an experimental analysis on the mental workload, demonstrating the potential of our method to reduce the cognitive burden on the user.






\bibliographystyle{IEEEtran}
\bibliography{bibliography}

\end{document}